\pdfoutput=1
\documentclass[journal]{IEEEtran}
\usepackage{cite}
\usepackage{amsmath,amssymb,amsfonts}
\usepackage[noend]{algpseudocode}
\usepackage{algorithmicx,algorithm}
\usepackage{graphicx}
\usepackage{textcomp}
\def\BibTeX{{\rm B\kern-.05em{\sc i\kern-.025em b}\kern-.08em
    T\kern-.1667em\lower.7ex\hbox{E}\kern-.125emX}}
\usepackage{graphics}
\usepackage{graphicx}
\graphicspath{{pic/}}
\usepackage{subfigure}
\usepackage[table]{xcolor}
\usepackage{multirow}
\usepackage{float}
\usepackage[section]{placeins}
\usepackage{booktabs}
\usepackage{threeparttable}

\usepackage{color, soul, framed}
\usepackage{multicol} 
\usepackage{pifont} 

\begin{document}

\title{Implicit-explicit Integrated Representations for Multi-view Video Compression}

\author{Chen Zhu,
        Guo Lu,~\IEEEmembership{Member,~IEEE},
        Bing He,
        Rong Xie,~\IEEEmembership{Member,~IEEE},
        Li Song,~\IEEEmembership{Senior~Member,~IEEE}

\thanks{
Chen Zhu, Guo Lu, Bing He, Rong Xie and Li Song are with the School of Electronic Information and Electrical Engineering, Shanghai Jiao Tong University, Shanghai, China (e-mail: zhuchenzc@sjtu.edu.cn; luguo2014@sjtu.edu.cn; sandwich\_theorem@sjtu.edu.cn; xierong@sjtu.edu.cn; song\_li@sjtu.edu.cn).

Li Song is also with the MoE Key Lab of Artificial Intelligence, AI Institute, Shanghai Jiao Tong University, China.

* The paper is under review.
}
}

\maketitle

\begin{abstract}

With the increasing consumption of 3D displays and virtual reality, multi-view video has become a promising format. 
However, its high resolution and multi-camera shooting result in a substantial increase in data volume, making storage and transmission a challenging task.
To tackle these difficulties, we propose an implicit-explicit integrated representation for multi-view video compression. 
Specifically, we first use the explicit representation-based 2D video codec to encode one of the source views. 
Subsequently, we propose employing the implicit neural representation (INR)-based codec to encode the remaining views. 
The implicit codec takes the time and view index of multi-view video as coordinate inputs and generates the corresponding implicit reconstruction frames.
To enhance the compressibility, we introduce a multi-level feature grid embedding and a fully convolutional architecture into the implicit codec. These components facilitate coordinate-feature and feature-RGB mapping, respectively.
To further enhance the reconstruction quality from the INR codec, we leverage the high-quality reconstructed frames from the explicit codec to achieve inter-view compensation. Finally, the compensated results are fused with the implicit reconstructions from the INR to obtain the final reconstructed frames.
Our proposed framework combines the strengths of both implicit neural representation and explicit 2D codec. Extensive experiments conducted on public datasets demonstrate that the proposed framework can achieve comparable or even superior performance to the latest multi-view video compression standard MIV~\cite{boyce2021mpeg} and other INR-based schemes in terms of view compression and scene modeling.
The source code can be found at https://github.com/zc-lynen/MV-IERV.

\end{abstract}


\section{Introduction}
Multi-view videos enable viewers to experience a three-dimensional space captured by camera arrays, offering a higher degree of freedom than 2D flat videos. With the rapid advancement of immersive video and 3D display technologies, multi-view videos have gained significant interest. However, it is very expensive to store and transmit the data of multiple camera views. Therefore, developing an effective way to encode multi-view videos has become a pressing need.

Over the past decade, the Moving Picture Experts Group (MPEG) has been committed to the development of multi-view video coding standards. Popular coding standards, such as 3D-HEVC~\cite{tech2015overview} and MIV~\cite{boyce2021mpeg}, rely on disparity to eliminate inter-view redundancy. Disparity compensation extracts matching points in different views to compute pixel-domain residuals, followed by the 2D video hybrid encoding process~\cite{bross2021overview, sullivan2012overview}.
However, these methods require high-precision depth maps and camera calibration parameters, and the corresponding hand-crafted modules are computationally complex, leading to poor coding efficiency and synthesis quality. 
Besides, to generate the immersive video, the rendering or synthesis procedure based on the multi-view videos is also challenging, hindering the development and practical applications.

\begin{figure}[t]
\centerline{\includegraphics[scale=0.63]{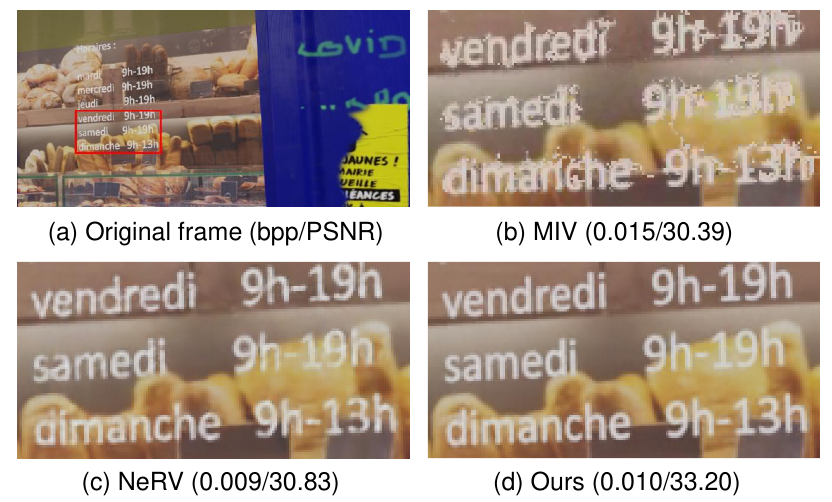}}
\caption{Visual quality of the reconstructed frames from different multi-view video compression systems. (a) is the original
frame. (b)-(d) are the reconstructed frames from MIV~\cite{boyce2021mpeg},
NeRV~\cite{chen2021nerv} and our method. 
Our proposed method consumes 0.010bpp while achieving the best objective quality (33.20) when measured by PSNR.}
\label{fig31}
\vspace{-5pt}
\end{figure}

Convolutional neural networks (CNNs) have shown remarkable success in image and video compression~\cite{balle2016end, liu2020deep, lu2019dvc, ma2019image}. However, deriving a learning-based multi-view video codec remains an open question, given the excessive data volume in multi-view format. A potential solution is the implicit neural representation (INR) paradigm~\cite{mildenhall2021nerf},
which trains a multilayer perceptron network (MLP) or deep model to fit the coordinate-texture mapping of a given scene, and the corresponding neural networks are quantized and entropy-coded for transmitting.
For example, 
ReRF~\cite{wang2023neural} presented neural radiance fields for motion and residual to exploit the inter-frame similarity in 3D scenes.
NeRV~\cite{chen2021nerv} proposed a frame-wise network to reconstruct 2D video frames given a time index input, achieving comparable rate-distortion (R-D) performance to commercial encoders. 
COIN~\cite{dupont2021coin} provided a pixel-wise MLP to represent the coordinate-color relationship of an image.
The main advantages of these methods are the simple architecture and native interpolation/rendering capability.
However, compared with the pixel-domain explicit representation based methods like CNN based approaches~\cite{hu2021fvc, lu2019dvc}, the reconstruction quality of existing INR  methods is not satisfactory due to the unstable backpropagation~\cite{colbrook2022difficulty}.

\begin{figure}[t]
\centerline{\includegraphics[scale=0.8]{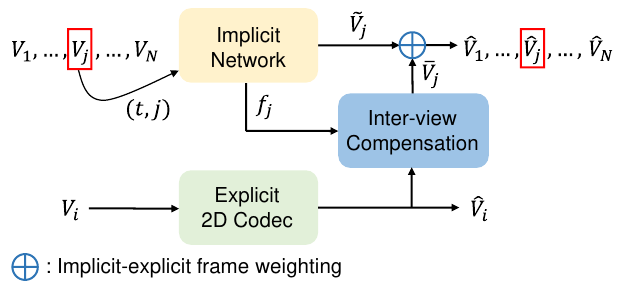}}
\caption{The proposed implicit-explicit integrated compression framework.
 2D codec compresses one of views into explicit reconstruction $\hat V_i$; For one other view $V_j$, implicit network is to generate the initial reconstruction $\widetilde V_j$; 
 Inter-view compensation warps the $\hat V_i$ by the flow information $f_j$ and produces the prediction view  $\overline V_j$ that is close to the current view; Implicit-explicit frame weighting fuses $\widetilde V_j$ and $\overline V_j$ to output the final reconstruction view $\hat V_j$.}
\label{fig1}
\end{figure}

To address these issues, 
we propose an implicit-explicit integrated framework for multi-view video compression that combines the advantages of these two different pipelines.
As given in Fig.~\ref{fig1}, firstly, one of the source views is compressed by existing explicit representation-based 2D codecs, such as HEVC~\cite{sullivan2012overview} and VVC~\cite{bross2021overview}. 
Then, for the remaining views, we propose to use a frame-wise implicit network to represent them by taking the frame index as input and producing the initial implicit reconstruction frame. 
This network intrinsically captures the content correlation between adjacent timestamps and views in the feature space.
To ensure efficient model training and inference, the implicit network utilizes a grid-based representation similar to~\cite{muller2022instant, li2023compressing, lee2022ffnerv} to model the feature space.
Concretely, we map a cascade of feature grids to the corresponding coordinate indexes and then pass the grid features through a series of upsampling and convolutional layers to produce high-resolution video frames.
In these convolutional layers, additional small feature grids are embedded to strengthen the spatial information.
The overall network architecture is suitable for compression and we utilize pruning, quantization, and entropy coding as network compression tools to reduce the model size. 
To further improve the prediction efficiency and reconstruction quality from the implicit codec, 
we propose performing explicit inter-view motion compensation by warping the high-quality reconstruction frame from the explicit 2D codec.
Eventually, we fuse the initial implicit reconstruction with the compensated results to obtain the final reconstructed frame.

We evaluate our proposed method on the MIV standard dataset~\cite{jung2020common} and D-NeRF dataset~\cite{pumarola2021d}. Experimental results show that our approach can achieve superior multi-view compression and modeling performance to the MIV standard test model and other 2D explicit or implicit representation methods~\cite{chen2021nerv, li2022nerv, lu2019dvc, chen2023hnerv, pumarola2021d, fang2022fast}. 
For instance, when compared to the MIV, the proposed method achieves a bitrate saving of approximately 37\% while maintaining the same objective reconstruction quality.

Our contributions can be summarized as follows:
\begin{itemize}
\item[$\bullet$]
We propose an implicit-explicit integrated representation for multi-view video compression. 
Our approach harnesses the strengths of both implicit representations and conventional explicit 2D codecs. 
\end{itemize}
\begin{itemize}
\item[$\bullet$]
We propose customized feature grids to support the training and compression of implicit convolutional networks.
We introduce an inter-view motion compensation mechanism, which exploits the high-quality reconstructed frame from explicit 2D codecs for better multi-view coding.
\end{itemize}
\begin{itemize}
\item[$\bullet$]
Experimental results show that our framework achieves comparable or even better multi-view compression performance compared to the immersive video encoder MIV and state-of-the-art INR-based methods.
\end{itemize}

\section{Related Work}
\subsection{Multi-view Video Compression}
For the early multi-view format video, binocular video, various compression methods have been proposed~\cite{dinstein1989compression,vetro2011overview} to reduce the view redundancy by utilizing disparity information.
The follow-up standard 3D-HEVC~\cite{tech2015overview} further improved the disparity estimation techniques and compressed the residual information using the 2D HEVC standard.
For multi-view video coding, the latest immersive video standard MIV~\cite{boyce2021mpeg} integrated multiple source views into a single 2D video that has eliminated inter-view redundancy.
This 2D video could be encoded by any flat video codec.
In summary, these hand-crafted compression methods heavily rely on complicated modules like disparity estimation or stitching, which limits their application in practical scenarios.

Recently, CNN-based binocular image and video coding methods have emerged~\cite{chen2022lsvc, deng2021deep, liu2019dsic}.
Liu~\emph{et al}. proposed DSIC, the first learning-based binocular image coding method that utilized a parametric skip function and conditional entropy model.
Chen~\emph{et al}. presented an end-to-end stereo video compression framework with feature-domain motion and disparity compensation.
However, these methods only consider stereo compression without considering the larger volume data compression introduced from the multi-view format. 
Hence, multi-view video compression is still an open and challenging problem.

\subsection{Implicit Neural Representations for Image and Video Compression}
Implicit neural representation is a novel approach to parameterize the signals~\cite{mildenhall2021nerf}. 
The fundamental concept behind INR is to use a neural network to approximate an object as a function that maps coordinates to their respective values. For example, a neural network can represent an image by taking the pixel coordinates as the inputs and producing the corresponding RGB values.
Several methods have suggested to generalize INR to a broader range of 3D shapes, including~\cite{liu2020neural, pumarola2021d, yu2021plenoctrees, tancik2022block, hong2022headnerf, muller2022instant,liu2022devrf, fang2022fast, li2023compressing, wang2023neural}.

In terms of image and video signals, 
Dupont~\emph{et al}. provided the image implicit neural representation COIN~\cite{dupont2021coin}, where a simple MLP was adopted to map spatial coordinates to colors.
In~\cite{yang2022tinc}, a tree-structured INR was proposed to adaptively represent different regions of an image.
Zhang~\emph{et al}.~\cite{zhang2021implicit} proposed INRs of motion and residual to capture the temporal variation in video sequences.
A scalable video INR was suggested in~\cite{kim2022scalable}, which memorized sparse positional features using latent grids.
Maiya~\emph{et al}.~\cite{maiya2022nirvana} proposed an autoregressive INR that modeled the correlations between co-located patches in each frame group.

Inspired by the effectiveness of INRs in representing visual signals, several methods have been proposed to achieve image or video compression based on INRs. 
For example, NeRV~\cite{chen2021nerv} introduced a frame-wise video compression method utilizing an MLP to transform the time index into spatial features. These spatial features were then upscaled to generate a complete frame.
By utilizing network compression tools such as pruning and quantization, 
NeRV demonstrated impressive results and attracted a lot of attention.
Subsequent studies such as~\cite{bai2022ps, chen2022cnerv, li2022nerv, lee2022ffnerv, zhao2023dnerv, he2023towards, chen2023hnerv} have further explored and improved upon NeRV. 
ENeRV~\cite{li2022nerv} optimized parameter chaining and convolution stacking,
while PS-NeRV~\cite{bai2022ps} adopted a patch-wise architecture to enhance network parallelism.
Zhao \emph{et al}.~\cite{zhao2023dnerv} introduced a network embedding of residual between adjacent frames.
In~\cite{he2023towards}, the optical flow was introduced in INRs to capture temporal information.
Meaningfully, there has been a consensus in some works that using MLPs for video compression was computationally and bitrate expensive,
thus they focused on exploring more compact network containers to build feature spaces.
In~\cite{lee2022ffnerv}, a temporal grid was employed as an alternative to MLP to improve the continuity of spatial features and the overall visual quality of compressed videos.
Works~\cite{chen2022cnerv}~\cite{chen2023hnerv} suggested using lightweight latent codes generated by auto-encoders as inputs to the implicit network backbone.
However, as noted in~\cite{colbrook2022difficulty}, it is difficult to achieve high-quality reconstruction solely relying on implicit network representations.
The rate-distortion performance of INR-based compression systems may not be satisfactory. This challenge becomes even more pronounced when applying INRs to multi-view video coding due to the larger data volume and increased complexity.

\section{MPEG Immersive Video Coding}

MPEG Immersive Video Coding (MIV)~\cite{boyce2021mpeg} is a state-of-the-art multi-view and immersive video compression technique that became an official standard in 2021.
It supports the viewing of volumetric contents with six degrees of freedom and provides specifications for both the encoding and decoding/rendering processes.

\begin{figure}[t]
\centerline{\includegraphics[scale=0.57]{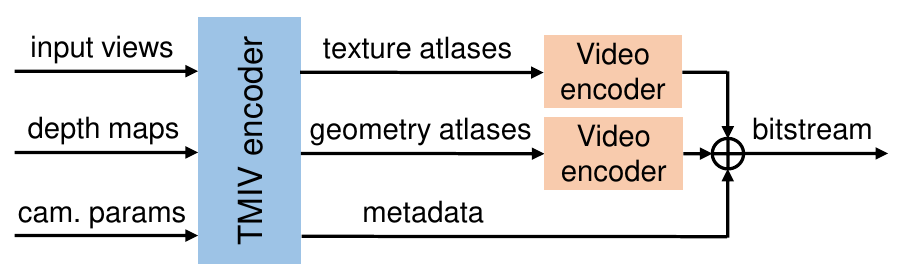}}
\caption{
The block diagram of MIV encoder.
}
\label{fig30}
\end{figure}

\textbf{MIV Encoder.}
As shown in Fig.~\ref{fig30}, The MIV encoder processes multiple source views along with corresponding geometry (depth) maps and camera parameters.
It outputs texture/geometry atlases and metadata.
The MIV encoding process consists of four main steps:
(1) View classification,
(2) Pixel pruning,
(3) View packing,
(4) 2D encoding.

View classification divides the source views into two categories: ``basic" views and ``additional" views. 
A ``basic" view has a high degree of content overlap with other views, its texture and geometry maps are fully packed into the atlas. 
In contrast, ``additional" views are predicted from the ``basic" view and pruned in the pixel domain to remove the inter-view redundancy, leaving only a small number of pixels in the atlas.

The pixel pruner takes into account the view hierarchy and projects either the ``basic" view or the pruned ``additional" view onto the current ``additional" view. 
It then locates and eliminates duplicate pixels in the projected view and the current view.
As illustrated in Fig.~\ref{fig31}(b), the gray area represents the pruned and eliminated pixels.

View packing involves merging the full ``basic" view and the pruned ``additional" views.
Its output is a flat atlas with a higher resolution than a single source view, as depicted in Fig.~\ref{fig31}(c).
This packing process is reversible,
therefore, the packing order, source view number and camera calibration parameters are included in the metadata, ensuring that the original views can be accurately recovered.

Following the above pipeline, the resulting atlases are encoded using any 2D codec, such as HEVC and VVC.
More MIV encoding details can be found in~\cite{boyce2021mpeg}.

\begin{figure}[t]
\centerline{\includegraphics[scale=0.55]{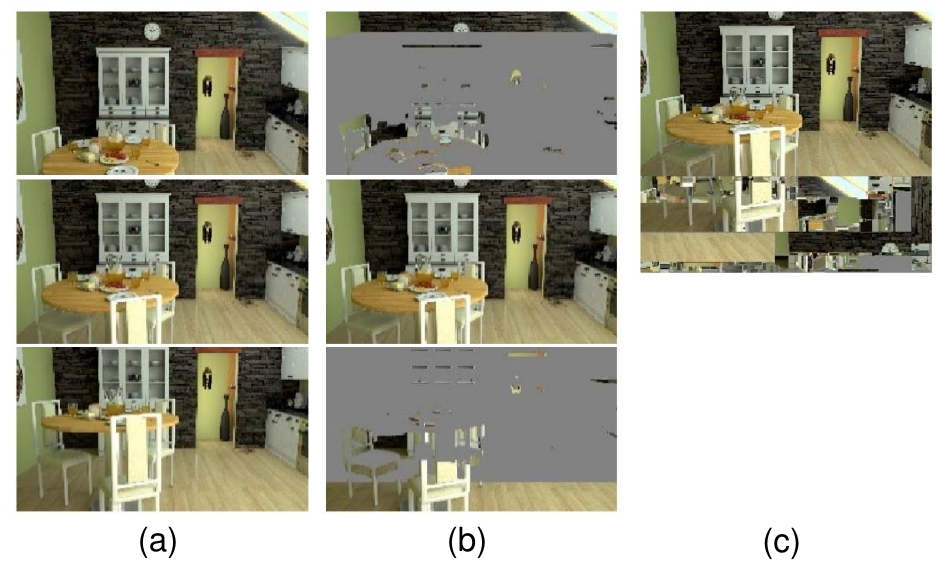}}
\caption{
The visualized process of MIV encoder.
(a) the texture maps of input views,
(b) the views after pixel pruning,
(c) the views after packing into atlas.
Similar operations are performed for the depth maps of input views.
}
\label{fig31}
\end{figure}

\begin{figure*}[t]
\centerline{\includegraphics[scale=0.65]{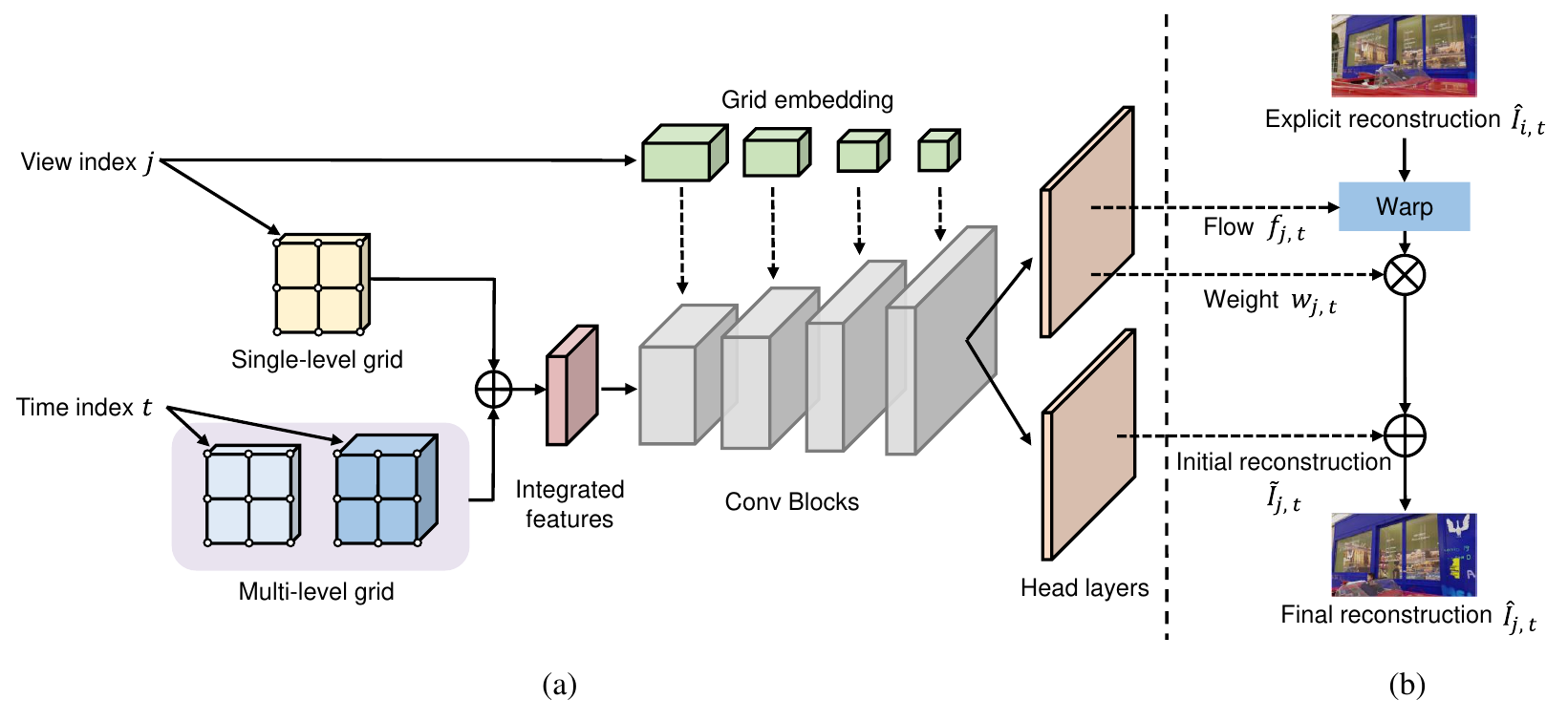}}
\caption{The workflow of the implicit network and the inter-view compensation. 
(a) Implicit network architecture. 
The temporal and perspective grid features are superimposed to input into the cascaded Convolutional (Conv) Blocks.
The grid embedding corresponding to the view input is utilized to modulate Conv Blocks.
(b) Inter-view compensation. It uses motion and weighting information learned from the implicit network to warp and aggregate the explicit reconstruction frame.}
\label{fig2}
\end{figure*}

\textbf{MIV Decoder/Renderer.}
Given the encoded atlases and metadata, the decoder performs the reverse process of the encoder to reconstruct the source views. Additionally, a crucial function of the decoder is to generate arbitrary intermediate views by utilizing reprojection and inpainting algorithms, based on the recovered source view information. 
In practical applications, based on the data provided by the video system, such as the position and orientation of the user's head, the MIV renderer will display the corresponding view content on the immersive device.

Recently, some studies~\cite{zhang2022deep, dziembowski2023immersive, garus2021immersive, lee2022efficient} have developed handcrafted or learning-based models to replace specific modules in the MIV framework.
However, a crucial problem remains unsolved.
The pruned and packed atlases exhibit noticeable spatial/temporal disorders,
which significantly hinders the effectiveness of 2D codec that relies on pixel correlations and leads to a certain degree of failure in coding modules such as intra/inter prediction.

Taking advantage of recent developments in implicit representation technology, where deep models internally estimate data correlations, we are inspired to design an implicit-explicit integrated inter-view prediction and encoding solution. 
We retain the view classification process, where one of the source views is selected as the ``basic" view and serves as a reference for inter-view prediction.
We simplify the cumbersome pruning and packing process in MIV and remove the algorithms based on depth maps, using only one implicit deep model to complete the compression of the remaining views.

In this article, we have not extensively improved the MIV renderer. Once the decoder obtains the reconstructed source views and their camera configurations by our proposed scheme, various excellent implicit representation schemes~\cite{pumarola2021d, liu2022devrf, fang2022fast, wang2023neural} can be employed to render the entire dynamic 3D scene without relying on depth maps, generating high-quality local perspectives. 



\section{Implicit-explicit Integrated Representations for Multi-view Videos}

The architecture of the proposed implicit-explicit integrated multi-view video compression is illustrated in Fig.~\ref{fig1}.
Let $\mathcal{M}={V_1, V_2, \dots, V_i, \dots, V_j, \dots, V_N}$ represent a multi-view video sequence, where each view $V_j$ is composed of multiple frames in a time series and $N$ is the number of views.
Firstly, one of the views $V_i \in \mathcal{M}$ is explicitly compressed using an existing 2D codec, such as the standard video codec HEVC, to generate a reconstructed video sequence $\hat{V}_i$.
For the remaining views, we use a neural network parameterized function $\mathcal{F}$ to implicitly represent them and ascertain their relevance, which will be optimized on the encoder side and sent to the decoder.
For example, to reconstruct each frame in the $j$-th view $V_j \in \mathcal{M}$, $j \neq i$, the view index $j$ and time index $t$ for each frame 
are fed into the function $\mathcal{F}$.
Then, we can obtain an initial reconstruction view $\widetilde{V}_j$.

To further exploit the correlations between different views, we propose an inter-view compensation module.
Based on the motion information $f_j$ from the implicit neural network $\mathcal{F}$, we warp the explicitly reconstructed video sequences $\hat{V}_i$ to the current view $V_j$, and the corresponding compensated results $\overline{V}_j$ are fused with the initial reconstruction view $\widetilde{V}_j$ to obtain the final reconstruction view $\hat{V}_j$.
For other views in this multi-view video sequence, we adopt the same compression pipeline.

In our proposed framework, we integrate both explicit and implicit representation-based codecs to achieve optimal multi-view video coding. In practical deployment, the bitstreams from explicit codecs and quantized implicit neural networks 
will be sent to the decoder side for reconstruction.

\subsection{View Selection and Explicit Compression}
We follow the depth map-based quality assessment process described by MIV to accomplish the view classification and selection process.
In particular, the depth map of the first frame of one specific view is projected onto the position of the remaining views. 
Then, a depth-level quality score of this view is calculated, where a lower deviation of the projected depth map from other views' depth maps corresponds to a higher score. In this paper, we designate the view with the highest quality score as the ``basic" view $V_i$, which is explicitly encoded utilizing 2D codec HEVC. 
The reconstructed view $\hat{V}_i$ will be employed to predict the other perspectives.

\subsection{Implicit Neural Network}



Here, we use $V_j$, $j \neq i$ as an example to demonstrate the detailed network architecture of the proposed implicit neural network in Fig.~\ref{fig2}.
Let $V_j = {I_{j,1}, I_{j,2},..., I_{j,t},..., I_{j,T}}$, where $I_{j,t}$ represents the $t$-th video frame in view $V_j$ and $T$ denotes the total timestamp.
In our proposed method, the implicit network inputs include the view index $j$ and time index $t$, and its learned mapping function $\mathcal{F}$ can be formulated as follows:
\begin{equation}
I_{j,t} = \mathcal{F}(j, t) \in \mathbb{R}^{H \times W \times 3}
\end{equation}
Here, $H \times W$ represents the resolution for the video frame.

\textbf{Feature Grid.} Given dual coordinate inputs, a common practice in previous implicit compression works~\cite{chen2021nerv, li2022nerv, bai2022ps} is to use Cosine functions and MLPs to modulate the coordinate values into 2D features and send features to subsequent network modules.
However, this approach makes it difficult to capture the abundant content change information presented in the time-view coordinate system of multi-view videos, such as motion and lighting.

Inspired by the recent developments in grid-based implicit representations~\cite{muller2022instant, li2023compressing, lee2022ffnerv}, 
we adopt multidimensional feature grids to process time-view coordinates.
We provide trainable grid tensors for the time index and view index, respectively.
The time grid consists of multiple temporal resolution levels, denoted as $g^{ti}=\{g^{ti}_1, g^{ti}_2, g^{ti}_3 \}, g^{ti}_k \in \mathbb{R}^{tr \times h \times w \times c }$, $k = 1, 2, 3$.
The view grid has a single resolution in the perspective dimension, expressed as $g^{vi} \in \mathbb{R}^{vr \times h \times w\times c }$.
Here, $tr$, $vr$, $h$, $w$ and $c$ represent the temporal resolution, perspective resolution, height, width, and channel number of the grid, respectively.
Then the temporal features $fea^{ti}$ and perspective features $fea^{vi}$ can be obtained through the following equations:
\begin{equation}
fea^{ti} = g^{ti}_1[t] + g^{ti}_2[\text{Floor}(t/2)] + g^{ti}_3[\text{Floor}(t/4)]
\label{eq11}
\end{equation}
\begin{equation}
fea^{vi} = g^{vi}[j]
\end{equation}
where Floor$(\cdot)$ denotes the floor operation.
In the experiments, $h$, $w$ and $c$ are adjustable parameters used to control the model bitrate, 
$vr$ equals to the total view number $N$, and the $tr$ of $g^{ti}_1, g^{ti}_2, g^{ti}_3$ is set to $T, T/2, T/4$, respectively.
The Equation~\ref{eq11} means that $g^{ti}_1$ is the time grid specific to each timestamp, 
whereas $g^{ti}_2$ and $g^{ti}_3$ are grids shared among multiple timestamps, with different sharing degrees.
Thus the multi-level temporal grid has the ability to capture spatial and short-term temporal features.

Furthermore, we employ a straightforward additive modulation strategy to merge $fea^{ti}$ and $fea^{vi}$, obtaining the integrated features $fea^{int}$ as follows:
\begin{equation}
fea^{int} = fea^{ti} + fea^{vi}
\label{equ:10}
\end{equation}

\begin{figure}[t]
\centerline{\includegraphics[scale=0.65]{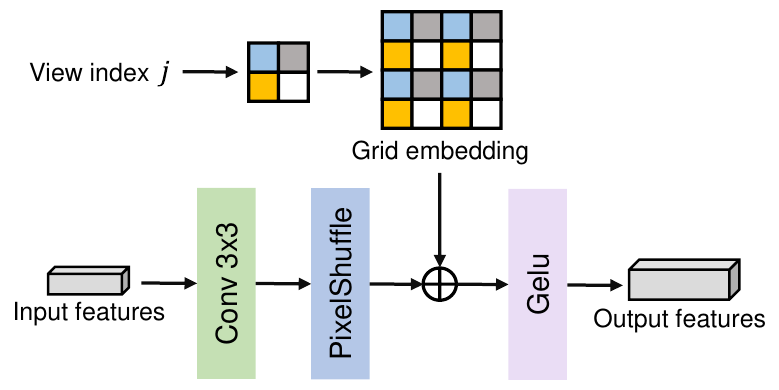}}
\caption{Architecture of the Conv Block, containing a Conv-Upsample-Activation pipeline. 
The kernels of upsampling and activation functions are Pixelshuffle and Gelu respectively.
The grid embedding contains the global features of the current view and modulates the Conv Block through additive operations.}
\label{fig3}
\vspace{-5pt}
\end{figure}

\textbf{Convolutional Structure.}
We design a convolutional neural network with multiple upsampling layers and grid embeddings to progressively scale the integrated time-view features $fea^{int}$ and generate the initial reconstruction frame $\widetilde{I}_{j,t}$. 
The data processing channel consists of five cascaded Convolutional (Conv) Blocks.
Each Conv Block follows the architecture shown in Fig.~\ref{fig3},
with the basic unit being a 3 $\times$ 3 convolutional layer, followed by a PixelShuffle upsampling layer and a Gelu activation function.

To further boost the representation capability of the implicit network, 
We introduce a trainable grid tensor $g^{em}_k$ in the $k$-th Conv Block.
$g^{em}_k$ has a shape of ${vr \times s_{cur} \times s_{cur} \times c_{cur}}$, 
where $s_{cur}$ and $c_{cur}$ represent the upscaling factor and channel number of the current Conv Block. 
To integrate the grid features into the basic unit, $g^{em}_k$ is copied multiple times and concatenated to form a grid embedding with a shape of $vr \times h_{cur} \times w_{cur} \times c_{cur}$, as depicted in Fig.~\ref{fig3}. Here, $h_{cur}$ and $w_{cur}$ denote the height and width of the current feature maps. 
The copied embedding is then added to the upsampled features before the Gelu activation function.
The above-described process can be formulated as follows:
\begin{equation}
\phi_{j, k} = \text{PixelShuffle}(\text{Conv}(fea_{j, k-1})) + \text{Copy}(g^{em}_k[j])
\end{equation}
\begin{equation}
fea_{j, k} = \text{Gelu}(\phi_{j, k})
\end{equation}
where $fea_{j, k}$ denotes the feature maps of the $k$-th Conv Block corresponding to the input view index $j$.  
The embedding $g^{em}_k$ captures the global features of frames within each view.
By continuously integrating the grid embeddings into the convolutional layers, the network can enhance the impact of coordinate information.

Finally, a $3 \times 3$ head convolutional layer is connected to the last Conv Block to generate the initial reconstruction $\widetilde{I}_{j,t}$.

\subsection{Inter-view Compensation} 
After applying the compression procedures described above, we obtain two reconstructed frames: an implicit reconstruction frame $\widetilde{I}_{j, t}$ and an explicit reconstruction frame $\hat{I}_{i, t}$ from the 2D codec.
While implicit models can inherently capture data correlations, such as contents from different views, their prediction efficiency is still inadequate.
To address this problem, we propose an inter-view compensation module in Fig.~\ref{fig2} based on the explicit reconstruction frame and the cross-view relationship.

Specifically, we input the output features of the last Conv block into another convolutional head layer and generate 2-channel motion information $f_{j,t}$ and 1-channel weight information $w_{j,t}$.
In the inter-view compensation module, we first use the motion information $f_{j,t}$ to perform cross-view motion compensation and warp the explicit reconstruction frame $\hat I_{i, t}$ to the current view.
Then, to address inaccuracies in the motion information, we use a convolutional layer to refine the warped frame and obtain a more accurate compensated frame $\bar{I}_{j, t}$.

Finally, based on the generated weight information $w_{j,t}$ from the implicit neural network, we combine the predicted frame $\bar{I}_{j, t}$ and the initial implicit reconstruction frame $\widetilde{I}_{j,t}$ in the following way:
\begin{equation}
\hat{I}_{j, t} = \widetilde{I}_{j, t} + w_{j, t} \odot \bar{I}_{j, t}  \\
\label{eq5}
\end{equation}
where $\hat{I}_{j, t}$ represents the final reconstruction frame.

\subsection{Training Loss}
To optimize the implicit neural network and the inter-view compensation module, we develop a 
joint loss function calculated over all the frames except the explicit compressed frames in the following way:
\begin{equation}
\begin{split}
L = \frac{1}{(N-1)T} \sum^{j \neq i}_{j=1} \sum^{T}_{t=1} \big[ \alpha || I_{j, t} - \hat I_{j, t} ||_{1}\\
 + (1 - \alpha) (1 - \text{SSIM}(I_{j, t}, \hat I_{j, t}))\big] \\
\end{split}
\label{eq5}
\end{equation}
where $|| \cdot ||_{1}$ and SSIM($\cdot$) denote L1 and SSIM loss functions respectively, $\alpha$ is the weight to balance losses.

\subsection{Model Compression}
For the implicit representation based compression, it is required to transmit the neural network model to the decoder side. 
Here, to reduce the size of the deep model,
three model compression techniques: pruning, quantization and entropy coding are employed to achieve a high compression ratio.

\textbf{Pruning.} Global unstructured pruning is executed to reduce the number of network parameters. We use the quantile function $=F^{-1}_{q}$ to adaptively select the threshold $\phi_{p}$ for pruning.
Network weights that are below the threshold will be omitted:
\begin{equation}
\phi_{p}=F^{-1}_{q}(\beta_{p})
\label{eq6}
\end{equation}
\begin{equation}
\phi_{i} = 
\left \{
\begin{aligned}
&\phi_{i}, && \text{if} \; \phi_{i} \geq \phi_{p} \\
&0, && \text{else}
\end{aligned}
\right.
\end{equation}
where $\beta_{p}$ denotes the proportion of parameters 
to be pruned.

\textbf{Quantization.}
We adopt a quantization-aware training method~\cite{zhou2016dorefa} to achieve numerical truncation operations.
In the quantization-aware training, the ``straight-through estimator" (STE)~\cite{bengio2013estimating} is used to solve the underivable problem of discrete functions, where the approximate gradient is passed to the convolutional layer during the backward pass.
For the forward strategy, we apply uniform quantization to both the model weights and the feature grids.
When quantizing a floating-point parameter $d$,  
we first normalize it to the range of [-1, 1] by applying the tanh($\cdot$) function.
Then, we scale the normalized value to obtain the quantized value $\hat d$:
\begin{equation}
\hat d = \text{sign}(d) \cdot \text{Floor}(|\text{tanh}(d)| \cdot Q_\text{max})
\end{equation}
where $Q_\text{max}$ is the maximum quantization value.
The signed 8-bit quantization is used in the implementation, correspondingly, $Q_\text{max}$ = $2^{8-1}-1$ = $127$. 


\textbf{Entropy coding.} As a common data coding technique, lossless Huffman Coding~\cite{huffman1952method} is adopted to further compress the quantized parameters and generate the bitstream of the implicit representation model.

\subsection{Bitstream}
Previous subsections explain the processing pipeline of the proposed compression framework, this subsection emphasizes the bitstream composition.
The proposed framework includes the following bitstream components:

(1) The bitstream of the ``basic" view, which is produced from the explicit standard encoder HEVC;

(2) The bitstream of other views, which is presented by the compressed feature grids and implicit network weights;

(3) Camera calibration parameters of all the source views, which include camera position and orientation. This metadata is encoded using the Exp-Golomb code, and the corresponding rate cost is negligible.

It is worth noting that the proposed framework does not require the transmission of any information related to the depth maps.


\section{Experiments}
\subsection{Experimental Setup}

\textbf{Datasets.} The public high-quality multi-view video datasets are few and we follow the common test conditions in the multi-view coding community. Specifically, we evaluate our proposed framework on seven different video sequences collected from the MIV standard testset~\cite{jung2020common}, each sequence has a camera number varied from 10 to 25 and is with a resolution of 1920$\times$1080. The first 100 frames of each view in the dataset are used for the comparisons between different baseline methods.
To further verify the effectiveness, we evaluate the perspective representation capabilities of the proposed method on the D-NeRF dataset~\cite{pumarola2021d}.
The D-NeRF dataset consists of eight scenes, each containing 100-200 views with a resolution of 800$\times$800.

\begin{figure*}[t]
  \centering
  \subfigure[R-D curves in terms of PSNR.]{\includegraphics[scale=0.53]{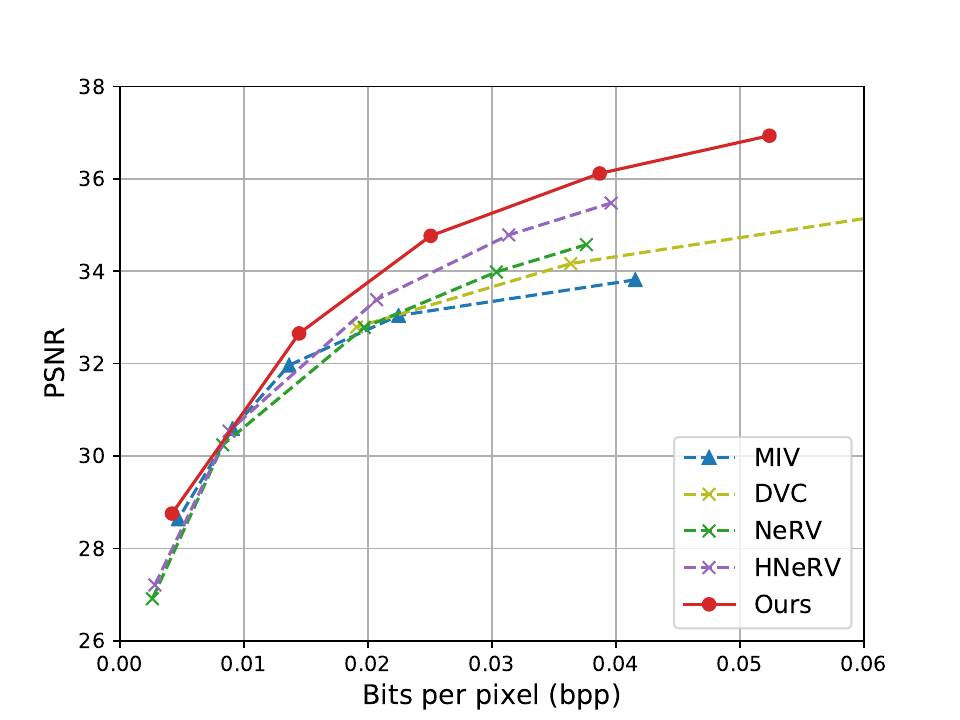}}
  \subfigure[R-D curves in terms of SSIM.]
  {\includegraphics[scale=0.53]{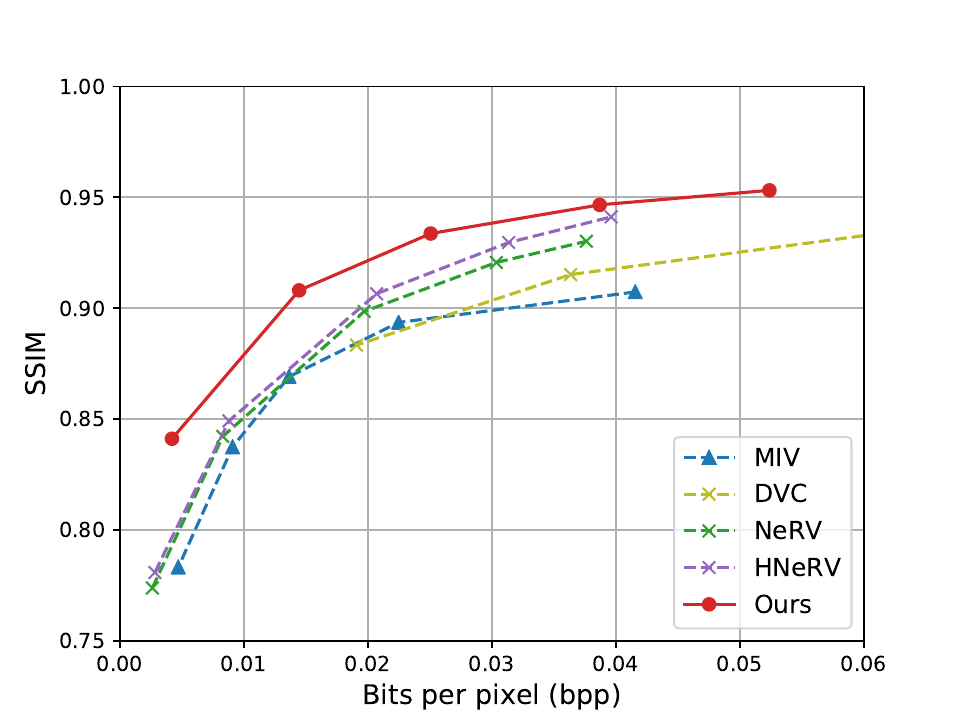}}
  \caption{Comparsion between the proposed method with the latest immersive video codec MIV and video codecs DVC, NeRV and HNeRV. }
  \label{fig4}
\vspace{-10pt}
\end{figure*}

\textbf{Implementation Details}.
In our design, the deep models, including the implicit network and inter-view compensation, are trained using the Adam optimizer~\cite{kingma2014adam} with parameters $\beta_1$=0.5, and $\beta_2$=0.999. 
The learning rate is set to 0.0005.
The network optimization is completed after 300\textit{th} epoch, and a batch size of 2 is used. 
For the 1080p videos in the MIV dataset, the $h$ and $w$ of multidimensional feature grids are set to 9 and 16, respectively. 
The upsampling factors of the five sequential Conv Blocks are 5, 3, 2, 2, 2 to generate the reconstructed pictures. 
For the 800$\times$800 videos in the D-NeRF dataset, $h$ and $w$ are set to 10 and 10. The upsampling factors are 5, 2, 2, 2, 2, respectively. 
The bitrate variation is achieved by adjusting the quantization parameter (QP) of the explicit codec and the channel number $c$ of feature grids. 
We employ the following one-to-one relationship between $c$ and QPs to compress the test sequences: $c_{i}$ $\rightarrow$ QP$_i$,
$c_{i}$ $\in \left\{20,30,40,60,80\right\}$ and 
QP$_{i}$ $\in \left\{43,38,33,28,23\right\}$,
$i$ = 1, 2, 3, 4, 5.
For the loss objective and the model pruning in Equation~\ref{eq5} and~\ref{eq6}, $\alpha$ and $\beta_{p}$ are set to 0.7 and 0.4, respectively.
Video quality is evaluated using the PSNR and SSIM~\cite{wang2004image} metrics and the compression ratio is indicated by bits-per-pixel (bpp).
We further calculate the BD-rate~\cite{bjontegaard2001calculation}, which reflects the average bit rate differences while maintaining the same quality, over the range of quality levels by a cubic polynomial fitting.
Negative values of BD-rate imply bit rate saving.
The experiments are conducted on NVIDIA RTX4090 GPUs.


\begin{table}[tbp]
\renewcommand\arraystretch{1.2}
\setlength\tabcolsep{10pt}
\caption{BD-rate results on multi-view compression task, with MIV as anchor.}
\begin{center}
\begin{tabular}{ c |  c  c }
\toprule[1.1pt]

Methods & BD-rate (PSNR) & BD-rate (SSIM)  \\
\midrule
DVC~\cite{lu2019dvc} & -12.2\% & -3.4\% \\
NeRV~\cite{chen2021nerv} & -0.5\% & -16.9\% \\
HNeRV~\cite{chen2023hnerv} & -10.4\% & -24.3\% \\
\midrule
Ours & -36.5\% & -48.2\% \\

\bottomrule[1.1pt]
\end{tabular}
\label{tab1}
\end{center}
\vspace{-10pt}
\end{table}

\begin{figure*}[t]
\centerline{\includegraphics[scale=0.55]{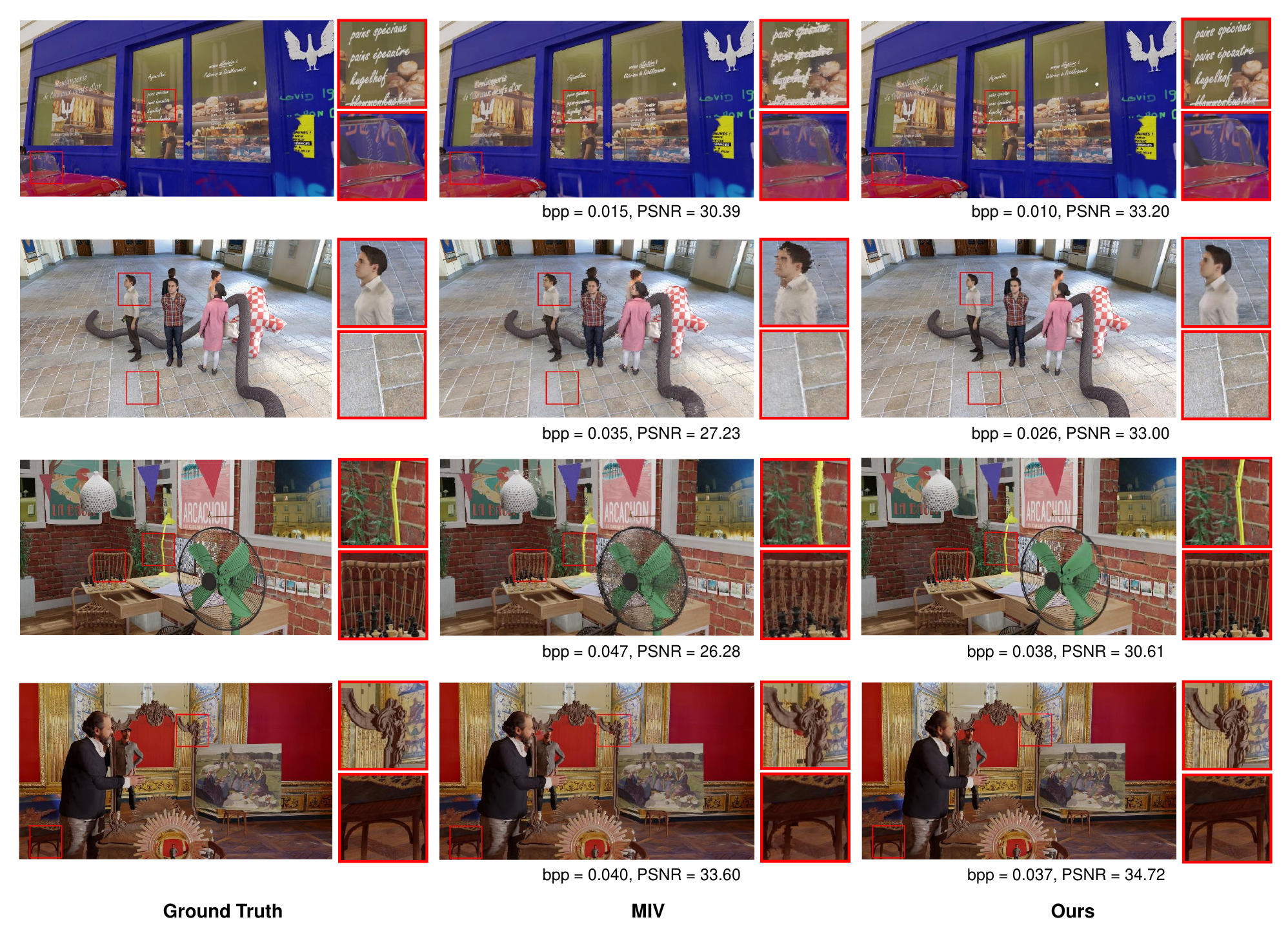}}
\caption{Qualitative comparison of different methods: ground truth video frames, reconstructions from MIV and our method. 
}
\label{fig5}
\end{figure*}

\textbf{Baseline Methods}.
In our experiments, we include the traditional multi-view video standard encoder MIV~\cite{boyce2021mpeg}.
It first eliminates the inter-view redundancy and then applies a flat video encoder to compress the pre-processed video.
For the 2D encoder in MIV and our method, we utilize the HEVC commercial encoder x265~\cite{guo2018compression} with the default $veryfast$ configuration.
To ensure a fair comparison, we do not consider the bitrate consumption of depth maps in MIV when assessing compression performance.
Additionally, we include the popular learning-based video codecs DVC~\cite{lu2019dvc}, NeRV~\cite{chen2021nerv} and HNeRV~\cite{chen2023hnerv} in our comparisons. 
It is important to note that these three video codecs independently encode/decode different views of a sequence, so the corresponding bitrate is the sum of the bits for all the views.
For DVC compression, we follow the suggested GOP size of 10 in the original article. 
As for the implicit compression of NeRV and HNeRV, we adopt the same configuration as proposed in~\cite{chen2021nerv}.

\subsection{Benchmark Results}
As the R-D curves and BD-rate results on the MIV dataset given in Fig.~\ref{fig4} and Table~\ref{tab1}, it can be found the performance of our method achieves similar or even better compression performance than the existing multi-view compression methods.
Take the MIV as the anchor method, our approach saves $36.5\%$ bitrate in terms of PSNR. 
For the SSIM metric, we obtain a $48.2\%$ compression performance gain compared with the anchor.
Interestingly, the performance gap between the proposed scheme and the anchor is widening at high bitrates. 
This is due to the application of model compression techniques, which eliminate a significant number of redundant implicit network parameters at these bitrate points.
Our approach also outperforms the learning-based 2D codec DVC and INR-based codecs NeRV/HNeRV. For example, DVC and NeRV only have 12.2\% and 0.5\% bitrate savings, respectively.




Fig.~\ref{fig5} shows some visualization results of the reconstructions.
It can be found at similar or even lower bitrates, our solution has overall higher image quality and better details in text, face and texture regions.
However, there are obvious edge effects and blurring artifacts caused by inaccurate depth estimation in the reconstruction of MIV.
One reason for the superior subjective performance of our method is that in the implicit-explicit hybrid framework, the trainable feature grids and the inter-view compensation module play a role in guiding the implicit network learning direction and generating high-quality frames.
In contrast, the MIV encoder usually fails to recover the pruned pixels correctly due to algorithmic errors such as depth estimation.

\begin{figure*}[t]
\centerline{\includegraphics[scale=0.68]{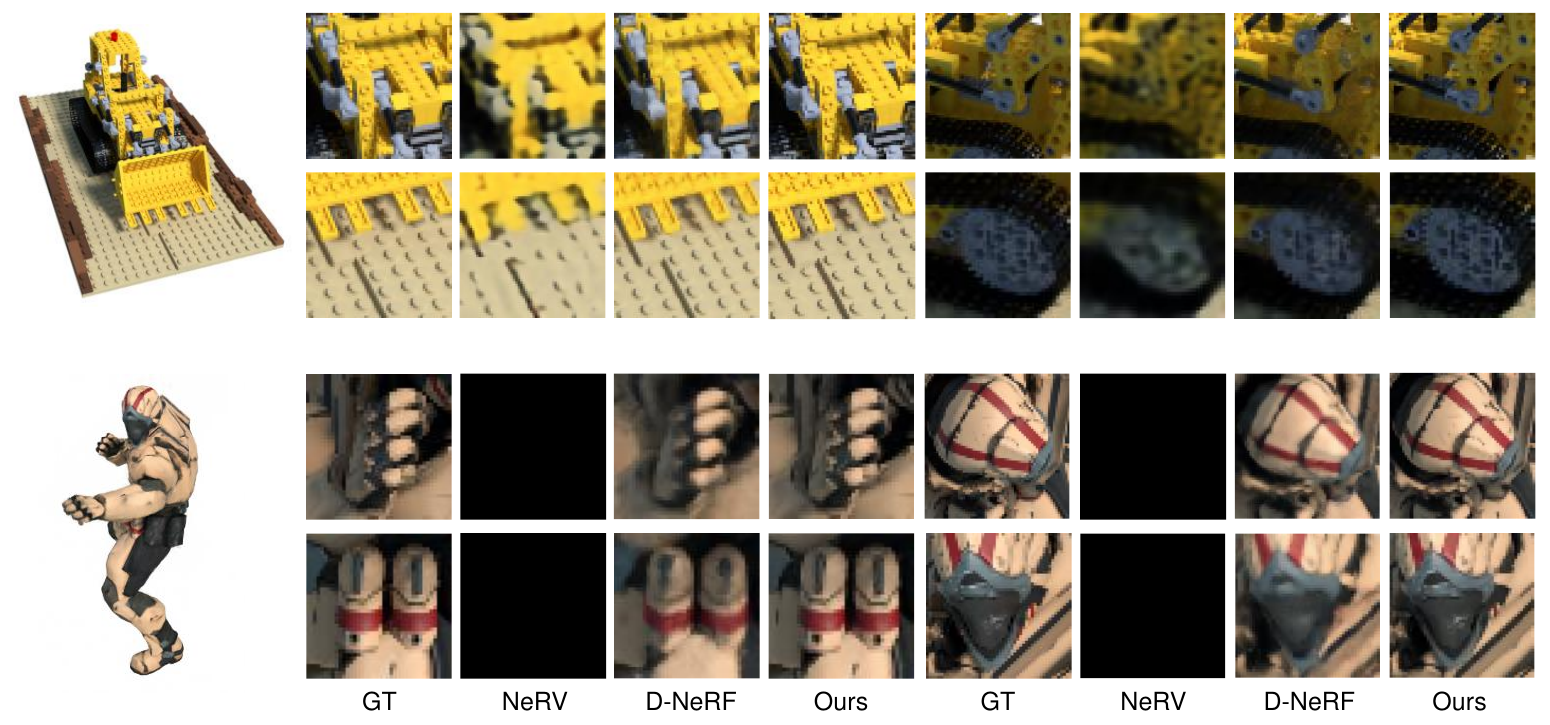}}
\caption{Qualitative comparison of different implicit representation methods.
From left to right: ground truth (GT), NeRV, D-NeRF and our proposed implicit model.
NeRV fails to produce any patterns on sequence $Hook$.}
\label{fig55}
\vspace{-10pt}
\end{figure*}

\begin{figure*}[t]
\centerline{\includegraphics[scale=0.57]{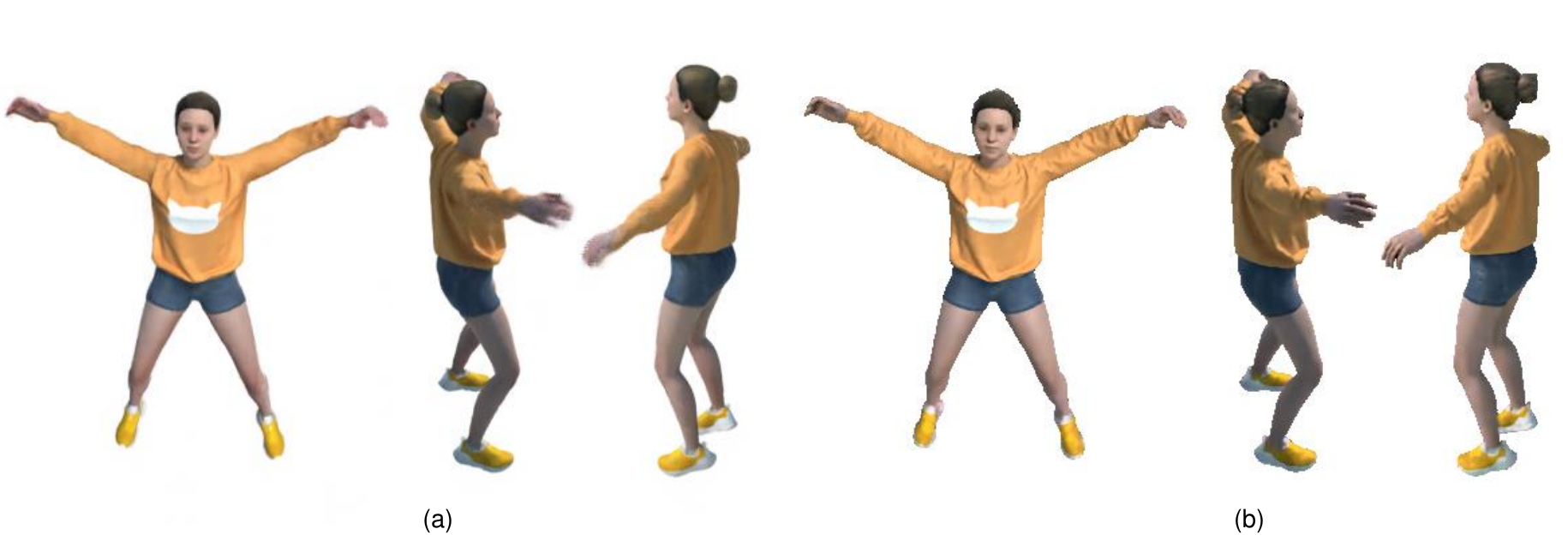}}
\caption{Qualitative comparison of different implicit representation methods.
(a): D-NeRF; (b): Our method. 
It can been seen that our approach has better subjective quality in the clothing, face, and body areas.}
\label{fig56}
\end{figure*}

\begin{table}[tbp]
\renewcommand\arraystretch{1.2}
\setlength\tabcolsep{5pt}
\caption{Compare with pixel-wise implicit representations. Params means the number of deep network parameters.}
\begin{center}
\begin{tabular}{ l | c c c c }
\toprule[1.1pt]

Methods & Params & PSNR $\uparrow$ & Training & Decoding  \\
        &        &                 &  FPS$\uparrow$ & FPS $\uparrow$  \\
\midrule
NeRV~\cite{chen2021nerv} &  4.0M & 26.8 & 0.11 & 115.4 \\
ENeRV~\cite{li2022nerv} &  3.9M & 30.1 & 0.06 & 67.8\\
D-NeRF~\cite{pumarola2021d} & 4.0M & 32.7 & 0.01 & 1.2 \\
Ours &  4.2M & 34.1 & 0.09 & 75.0 \\
\midrule
NeRV~\cite{chen2021nerv} & 8.1M & 30.7 & 0.09 & 93.7 \\
ENeRV~\cite{li2022nerv} &  7.9M & 33.8 & 0.05 & 50.3\\
TiNeuVox~\cite{fang2022fast} & 8.0M & 33.1 & 2.53 & 253.2 \\
Ours &  8.1M & 36.7 & 0.08 & 72.6 \\
\bottomrule[1.1pt]

\end{tabular}
\label{tab2}
\end{center}
\vspace{-10pt}
\end{table}

\subsection{Implicit Model Representation}
We conduct comparison experiments to evaluate the representation ability of our proposed implicit network with frame-wise implicit representations NeRV~\cite{chen2021nerv} and ENeRV~\cite{li2022nerv}. 
Additionally, we compare our model with 3D implicit representations D-NeRF~\cite{pumarola2021d} and TiNeuVox~\cite{fang2022fast}.
In D-NeRF and TiNeuVox, coordinate deformation was implicitly modeled to enhance temporal information.
For a fair and convenient comparison, we train two different-sized models and optimize all the representations with the same number of epochs.
In the comparison, we exclude the proposed motion compensation module and rely solely on the implicit model to fit all the source views.

We conduct tests on the D-NeRF dataset and the results, including PSNR, training speed, and decoding time, are summarized in Table~\ref{tab2}.
Notably, our method achieves PSNR improvements of approximately 7dB and 6dB compared to NeRV, at model sizes of 4M and 8M respectively. 
Additionally, our method incurs a modest decrease in the codec speed, with training and decoding FPS (Frame Per Second)  at approximately 0.1 and 70.
One significant advantage of our method lies in the grid-based representation. By introducing feature grids, the implicit network is able to capture rich temporal and perspective information while minimizing the number of network parameters. 
This allows the network to adapt to multi-view scenarios with arbitrary views and timestamps, and enables fast convergence aided by feature priors.

Furthermore, when compared to the 3D representations of D-NeRF (4M) and TiNeuVox (8M), our approach also achieves 1.4dB and 3.6dB PSNR improvements, respectively. This demonstrates that these scene rendering models have a surplus of unnecessary network parameters when fitting a limited number of perspectives.

To further verify the representation capacity of our model, we provide a qualitative comparison in Fig.~\ref{fig55} and Fig.~\ref{fig56}. The results clearly show that our approach generates more abundant texture details compared to other implicit representations, while also mitigating blurring artifacts.



\subsection{Ablation Studies}
Network component analysis for multidimensional feature grids (Grid-Fea), grid embedding (Grid-Emb), inter-view compensation (IVC) and model compression techniques, are provided in this subsection. 
The module switch results are shown in Table~\ref{tab3},
which reflects that disabling these proposals leads to quality decline compared to the original model.
Table~\ref{tab5} provides the superimposed test results of model compression tools.

\textbf{Multidimensional Feature grid.}
In our study, we adopt time grid (Grid-Fea-T) $g^{ti}$ and view grid (Grid-Fea-v) $g^{vi}$ to capture the characteristics of multi-view scenes.
We conduct experiments by removing the time grid and view grid individually, using a globally shared grid tensor $g^{glo} \in \mathbb{R}^{1 \times h \times w \times c}$
to learn the input features of upsampling layers. 
The experimental results in Table~\ref{tab3} for the alternative methods indicate that the PSNR without these two grids is degraded by 6.6dB and 5.9dB at the model size of 8M, respectively. 
Additionally, turning off these two grids results in an 8.2dB PSNR drop. Hence, accurate spatio-temporal context prior is crucial for frame-wise reconstruction.

\begin{table}[t]
\renewcommand\arraystretch{1.2}
\setlength\tabcolsep{10pt}
\caption{Ablation tests on the proposed modules for multi-view video representation.}
\begin{center}
\begin{tabular}{l | c | c  c }
\toprule[1.1pt]

Models & Params & PSNR $\uparrow$ & SSIM  $\uparrow$ \\
\midrule
Original & 8.0M & 37.6 & 0.952 \\
w/o Grid-Fea & 7.7M & 29.4 & 0.868  \\
w/o Grid-Fea-T & 7.8M & 31.0 & 0.923  \\
w/o Grid-Fea-V & 7.8M & 31.7 & 0.874  \\
w/o Grid-Emb & 7.9M & 37.2 & 0.948 \\
w/o IVC & 7.9M & 36.7 & 0.945 \\
\midrule
Original & 12.2M & 39.3 & 0.964 \\
w/o Grid-Fea & 11.8M & 30.0 & 0.865\\
w/o Grid-Fea-T & 12.0M & 32.0 & 0.937  \\
w/o Grid-Fea-V & 12.0M & 33.4 & 0.887  \\
w/o Grid-Emb & 12.1M & 39.1 & 0.963   \\
w/o IVC & 12.1M & 38.3 & 0.959 \\

\bottomrule[1.1pt]
\end{tabular}
\label{tab3}
\end{center}
\vspace{-5pt}
\end{table}

\begin{table}[t]
\renewcommand\arraystretch{1.2}
\setlength\tabcolsep{8pt}
\caption{PSNR / compression ratio results of three introduced model compression techniques: pruning (P),  quantization (Q), entropy coding (E).}
\begin{center}
\begin{tabular}{l | c  c | c }

\toprule[1.1pt]
Params &  4M & 8M & avg. \\
\midrule
Original & 35.07/1.00 & 37.8/1.00 & 36.43/1.00 \\
\midrule
Q & 34.19/4.00 & 36.86/4.00 & 35.53/4.00 \\
Q+P & 33.42/6.74 & 36.02/6.62 & 34.72/6.68 \\
Q+P+E & 33.42/6.77 & 36.02/7.63 & 34.72/7.18 \\
\bottomrule[1.1pt]

\end{tabular}
\label{tab5}
\end{center}
\vspace{-5pt}
\end{table}

\textbf{Grid Embedding.}
In this paper, grid embedding $g^{em}_k$ is utilized to fuse the features of input views and the intermediate convolution features. 
The experimental results of the alternative approach denoted by w/o Grid-Emb in Table~\ref{tab3} show that the PSNR without grid embedding drops by 0.4dB and 0.2dB at model sizes of 8M and 12M, respectively. This demonstrates that enhancing perspective information in the network can improve the final multi-view reconstruction effect.

\begin{figure}[t]
\centerline{\includegraphics[scale=0.7]{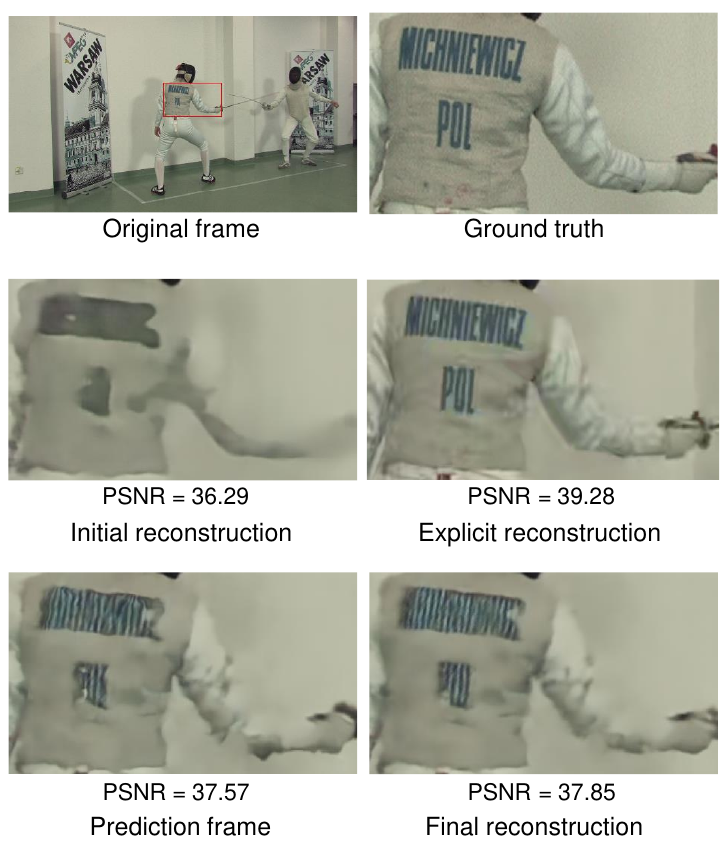}}
\caption{Qualitative comparison for inter-view compensation module.}
\vspace{-5pt}
\label{fig9}
\end{figure}

\textbf{Inter-View Compensation.}
To demonstrate the effectiveness of the proposed inter-view compensation module, we further proposed an alternative solution by training each view and each frame independently without aggregating the explicit reconstructed frame.
Results in Table~\ref{tab3} show this approach leads to a 0.9dB/1.0dB PSNR degeneration at 8M/12M models.
It demonstrates that providing high-quality frames as auxiliary information of the network is beneficial to model convergence and performance improvement.
We also visualize the compensation results in Fig.~\ref{fig9}, it can be seen that the prediction frame, obtained by explicit reconstruction based compensation, is clearer than the initial reconstruction, 
the subsequent fusion process of compensated prediction frame and initial reconstruction in the framework leads to a high-quality final reconstruction frame.


\begin{table}[t]
\renewcommand\arraystretch{1.2}
\setlength\tabcolsep{10pt}
\caption{Encoding and decoding speed compared with state-of-the-art codecs are reported.}
\begin{center}
\begin{tabular}{l | c  c }
\toprule[1.1pt]

Methods & Encoding & Decoding \\
 & FPS $\uparrow$ &  FPS $\uparrow$ \\
\midrule
MIV~\cite{boyce2021mpeg} & 0.23 & 2.7 \\
DVC~\cite{lu2019dvc} & 0.04 & 15.2 \\
NeRV~\cite{chen2021nerv} & 0.09 & 93.7 \\
HeRV~\cite{chen2023hnerv} & 0.05 & 87.8 \\
\midrule
Ours & 0.08 & 72.6 \\

\bottomrule[1.1pt]
\end{tabular}
\label{tab4}
\end{center}
\vspace{-10pt}
\end{table}

\textbf{Model Compression Strategy.}
In the proposed method, we use several model compression techniques to reduce the model and achieve a variable compression bitrate.
Specifically, network pruning reduces the number of model parameters, the quantization technique decreases the bit number occupied by parameters, and entropy coding removes symbol redundancy.
The detailed results are shown in Table~\ref{tab5}, it is observed that 
through the superimposed test on two different original network sizes, 
the quantization can bring 4.0 times the compression ratio with video quality degradation of 0.9dB PSNR. 
When the pruning tool is added, the compression ratio varies from 4.0 to 6.7 means that this technique can save nearly 45\% of the bitrate cost, and the corresponding quality loss is controllable.
Huffman Coding reduces the model size by 1.1 times, without any influence on the reconstruction quality.
In the future, other advanced model compression algorithms, such as distillation~\cite{neill2020overview}, can be applied to further improve the compression performance.

\subsection{Complexity Analysis}
We compare encoding and decoding speeds with state-of-the-art codecs.
In our implementation, all the learning-based schemes are conducted with GPU acceleration, other approaches are evaluated on the multi-core CPU platform.
As illustrated in Table~\ref{tab4},
since INR-based methods need online training on the encoder side, the encoding time of our method is slower than that of the hand-crafted encoder MIV.
However, since the entire framework is lightweight, the results show that the decoding speed of our method is nearly 70 FPS, which is close to previous implicit video codecs NeRV and HNeRV.

\section{Discussion}

\textbf{Limitation.}
As an attempt to achieve multi-view video compression with implicit radiance field modeling, our scheme still faces some challenges. Firstly, fast-moving objects like the electric fan in Fig.~\ref{fig5} can lead to blurry artifacts in the reconstruction results. 
To enhance the prediction effect, it may be necessary to incorporate fine temporal optical flow estimation in the implicit model training. Secondly, while we have employed a grid-based implicit feature extractor with fast fitting speed, the encoding and decoding time falls short of meeting the requirements for practical applications. This presents a significant area for future research.

\textbf{Conclusion.}
In this paper, we have proposed an implicit-explicit integrated framework for multi-view video compression. 
The proposed framework inherits the advantages of both traditional explicit codecs and advanced implicit neural representations.
The implicit network is iterated with the guidance of explicit reconstruction frames.
Customized feature grids are introduced to enhance the representation ability of implicit networks. 
Multiple model compression techniques are employed to minimize the bit cost of network weights.
As experimental results show, the proposed approach outperforms the existing depth-based multi-view video encoder and frame-wise implicit compression methods at multi-view compression and representation tasks.
Our work provides a promising scheme for future immersive video coding, 
efficient network modules and compression techniques can be integrated into the proposed framework in future works.
    
{\small
\bibliographystyle{ieee_fullname}
\bibliography{egbib}
}

\end{document}